\documentclass[10pt,twocolumn,letterpaper]{article}

\usepackage[pagenumbers]{wacv} 

\usepackage{graphicx}
\usepackage{amsmath}
\usepackage{amssymb}
\usepackage{booktabs}
\usepackage{multicol}
\usepackage{multirow}
\usepackage[table,xcdraw]{xcolor}
\usepackage{amsfonts}

\usepackage[pagebackref,breaklinks,colorlinks]{hyperref}

\usepackage[capitalize]{cleveref}
\DeclareMathOperator*{\argmin}{argmin}
\crefname{section}{Sec.}{Secs.}
\Crefname{section}{Section}{Sections}
\Crefname{table}{Table}{Tables}
\crefname{table}{Tab.}{Tabs.}

\def\wacvPaperID{12} 
\def\confName{WACV}
\def\confYear{2025}

\title{Mahalanobis k-NN: A Statistical Lens for Robust Point-Cloud Registrations}
\author{Tejas Anvekar, Shivanand Venkanna Sheshappanavar  \\ Geometric Intelligence Research Lab \\ University of Wyoming, 1000 E University Ave, Laramie, WY 82071}

\begin{document}
\twocolumn[{%
\renewcommand\twocolumn[1][]{#1}%
\maketitle
\begin{center}
    \centering
    \captionsetup{type=figure}
    \includegraphics[width=0.95\linewidth]{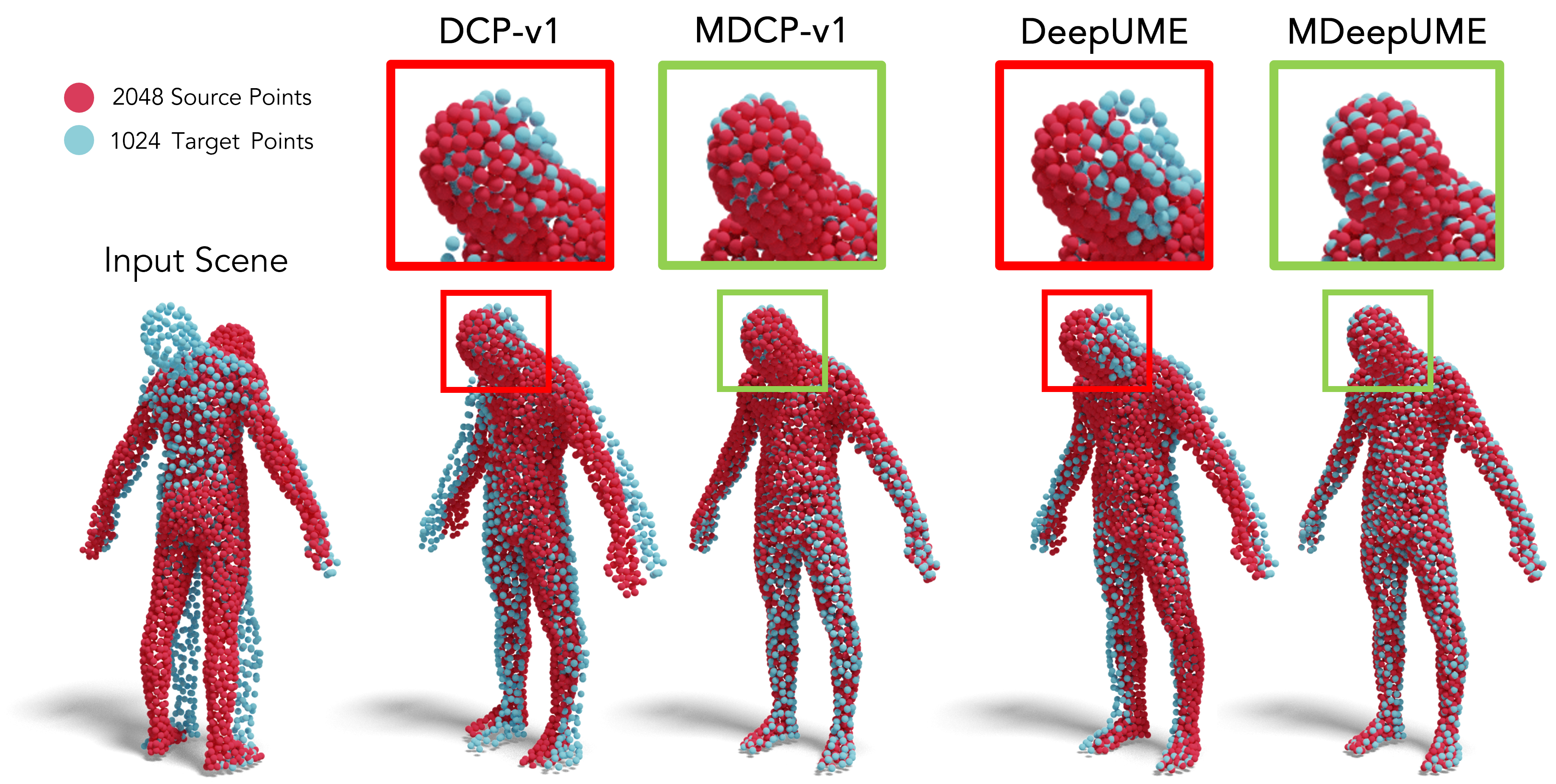}
    \captionof{figure}{The visual supremacy of our proposed methodologies, MDCP-v1 and MDeepUME, becomes apparent in a point cloud registration,  involving a target point cloud with only half the points compared to the source point cloud. In the visualization, regions highlighted in \textcolor{red}{red} illustrate the limited performance of DCP-v1\cite{DCP} and DeepUME\cite{DeepUME}, while the regions highlighted in \textcolor{green}{green} demonstrate the resilience of the proposed Mahalanobis versions of DCP-v1\cite{DCP} and DeepUME\cite{DeepUME}.}
    \label{fig:teaser}
\end{center}
}]
\maketitle


\begin{abstract}
In this paper, we discuss Mahalanobis k-NN: A Statistical Lens designed to address the challenges of feature matching in learning-based point cloud registration when confronted with an arbitrary density of point clouds. We tackle this by adopting Mahalanobis k-NN's inherent property to capture the distribution of the local neighborhood and surficial geometry. Our method can be seamlessly integrated into any local-graph-based point cloud analysis method. In this paper, we focus on two distinct methodologies: Deep Closest Point (DCP) and Deep Universal Manifold Embedding (DeepUME). Our extensive benchmarking on the ModelNet40 and FAUST datasets highlights the efficacy of the proposed method in point cloud registration tasks. Moreover, we establish for the first time that the features acquired through point cloud registration inherently can possess discriminative capabilities. This is evident by a substantial improvement of about 20\% in the average accuracy observed in the point cloud few-shot classification task, benchmarked on ModelNet40 and ScanObjectNN. 


\end{abstract}


\section{Introduction}
\label{sec:intro}
With the rapid progress of high-precision sensors like LiDAR~\cite{lidar} and Kinect~\cite{kinect}, point clouds are prevalent data format for 3D representation. Due to sensor limitations in capturing only partial views, registration algorithms are crucial for amalgamating scans into comprehensive 3D scenes. Point cloud registration involves determining transformation matrices between point cloud pairs, facilitating the fusion of partial scans to produce coherent 3D representations.

Registration is a fundamental task spanning various computational fields such as medical imaging~\cite{medicalImaging}, robotics~\cite{robotics}, autonomous driving~\cite{autoDriving}, and computational chemistry~\cite{ComChe}. Estimating transformation matrices serves applications like 3D reconstruction~\cite{reconstruction} and precise 3D localization~\cite{Localization}. It underpins the development of intricate 3D maps for autonomous driving~\cite{autoDriving}, environment reconstruction in robotics~\cite{robotics}, and improved safety in mining. Additionally, registration's ability to facilitate high-precision localization proves invaluable for entities like driverless cars~\cite{autoDriving}, ensuring precise positioning and interaction within the 3D environment.

The registration of point clouds has garnered significant attention through closed-form~\cite{ICP,goICP,FGR,aligningPointCloudView,ume} and learning-based~\cite{pointnetLK,DCP,DeepGMR,DeepUME,rgm} methodologies. DCP~\cite{DCP} tackles the challenge of feature matching (correspondences) by initially estimating local features via weight-shared DGCNN~\cite{dgcnn} (Dynamic Graph Convolutional Neural Network). Subsequent steps involve an attention-based module and differentiable SVD modules for point-to-point registration. Conversely, DeepUME~\cite{DeepUME} adopts a distinct approach by projecting points into a space invariant under \(\mathcal{SO}(3)\) transformations. This space is leveraged to compute invariant per-point features that address real-world data sampling issues. Employing two weight-shared transformers and DGCNN~\cite{dgcnn}, correspondences are established and fed into UME~\cite{ume} for rigid transformation parameter estimation. A limitation of both DCP~\cite{DCP} and DeepUME~\cite{DeepUME} is susceptibility to failure in scenarios where either the source or target point cloud exhibits arbitrary (lower) density than the others—an occurrence commonly encountered in 3D scan registration from diverse sensors. This limitation is attributed to the methods' reliance on estimating per-point correspondences using local features extracted from edge-conv~\cite{dgcnn}, which operates on graphs constructed using k-nearest neighbors (k-NN). 

We introduce Mahalanobis k-NN: A Statistical Lens for Robust Point-Cloud Registration to mitigate aforementioned issue. The efficacy of Mahalanobis distance as an evaluation metric has already been demonstrated in reference-based point cloud quality assessment~\cite{MD1}. Our proposed Mahalanobis k-NN can be used as a plugin for any point-cloud registration method. In this paper, we propose mahalnobis versions of DCP-(v1,v2)~\cite{DCP} and DeepUME~\cite{DeepUME}. Both methods operate on diverse point-cloud representation spaces. This augmentation facilitates pointcloud registration as shown in Figure \ref{fig:teaser}; across various publicly available 3D datasets, encompassing diverse benchmarking scenarios, including: 1) unseen-category evaluation, 2) robustness to various noise types, and 3) efficiency towards varying point densities. Additionally, we demonstrate the discriminative capabilities of the proposed Mahalanobis lens through point cloud few-shot classification tasks. In this context, models pre-trained for registration tasks leverage DGCNN~\cite{dgcnn} features for few-shot evaluation on both ModelNet40~\cite{modelnet} and ScanObjectNN~\cite{SONN} datasets. Finally, we summarize our contributions as follows:

\begin{itemize}
    \item We propose Mahalanobis k-NN: A Statistical Lens for Robust Point-Cloud Registration by incorporating Mahalanobis distance in:
    \begin{itemize}
        \item Deep Closest Point (DCP)~\cite{DCP} that operates on Euclidean coordinate space.
        \item Deep Universal Manifold Embedding (DeepUME)~\cite{DeepUME} that operates on \(\mathcal{SO}(3)\) invariant space.
    \end{itemize}
    \item We demonstrate the efficacy of the proposed method on various point cloud registration tasks on a variety of publicly available data sets and compare the results with state-of-the-art techniques.  
    \item We perform an endurance test to evaluate the robustness and generalization of the proposed method. We achieve state-of-the-art results compared to other point cloud registration methods on all benchmarking strategies and on almost all evaluation metrics.
    \item To the best of our knowledge, we are the first to demonstrate discriminative prowess inherited in point cloud registration task by benchmarking point cloud few-shot classification on ModelNet40~\cite{modelnet} and ScanObjectNN~\cite{SONN} datasets through features extracted by DCP~\cite{DCP}, we report up to an average of 20\% increment when Mahalanobis is incorporated. 
\end{itemize}

\section{Related Works}



\noindent \textbf{Optimization Based.} Preceding the advent of the deep learning era in 3D point cloud registration, one common strategy involved extracting and matching spatially local features, as demonstrated in studies like~\cite{comprehensiveEvalFeatureDescriptor,spinImages,aligningPointCloudView,RepeatabilityRobust3Dmatching,fastRobustDescriptor,rotationalContour}. Many existing methods in this category are adaptations of 2D image processing solutions, such as variants of 3D-SIFT~\cite{3Dsift} and the 3D Harris key-point detector~\cite{harris3D}. However, key-point matching in 3D presents challenges due to the absence of a regular sampling grid, artifacts, and sampling noise, leading to high outlier rates and localization errors. To achieve global alignment, key-point matching typically utilizes outlier rejection methods like RANSAC~\cite{ransac} and is refined using local optimization algorithms~\cite{ICP,scanRegistration3Dndt,registrationWithoutICP,iterativePointMatching}. Notably, DGR~\cite{DGR} follows a similar paradigm, but it incorporates learnable inlier detection. Researchers have proposed numerous works for handling outliers and noise~\cite{trimmedICP}, formulating robust minimizers~\cite{robustRegistration2D3D}, and devising more suitable distance metrics. The widely used Iterative Closest Point algorithm (ICP)~\cite{ICP,iterativePointMatching}, a popular refinement algorithm, constructs point correspondences based on spatial proximity and employs a transformation estimation step. Over time, various ICP variants~\cite{efficientICP,goICP,sparseICP} have been proposed to enhance convergence rate, robustness, and accuracy in the 3D point cloud registration.\\

\noindent \textbf{Learning Based}. Apart from optimization-based methods, registration approaches have also leveraged learning-based techniques. Pioneered by PointNet ~\cite{pointnet} and further developed by DGCNN~\cite{dgcnn}, learning from data in a task-specific manner has become a powerful tool in point cloud registration. These learned point cloud representations can be exploited for robust registration, as evidenced in studies like~\cite{pointnet++,DCP,prnet}. For instance, PointNetLK ~\cite{pointnetLK} minimizes learned feature distance using a differentiable Lucas-Kanade algorithm ~\cite{LK}, while DCP~\cite{DCP} addresses feature matching through attention-based modules and differentiable SVD modules for point-to-point registration. Recently,~\cite{rgm} introduces a transformation of point clouds into graphs and utilizes deep graph matching to extract deep feature soft correspondence matrices. To overcome this limitation, researchers have proposed approaches~\cite{DeepUME} employing \(\mathcal{SO}(3)\) invariant coordinate systems to provide a global registration solution. Another method, DeepGMR ~\cite{DeepGMR}, deals with pose-invariant correspondences between raw point clouds and Gaussian mixture model (GMM) parameters, recovering transformations from the matched mixtures, although it exhibits performance degradation in the presence of sampling noise. Moreover, DeepUME~\cite{DeepUME} lies in its innovative fusion of the closed-form Universal Manifold Embedding (UME) method and a deep neural network to address the challenge of registering sparsely and unevenly sampled point clouds subjected to large transformations. This unified framework, trained end-to-end and without supervision, employs an \(\mathcal{SO}(3)\)-invariant coordinate system to learn joint-resampling strategies and invariant features. 

While DCP~\cite{DCP} and DeepUME~\cite{DeepUME} exhibit impressive performance across various real-world scenarios in point cloud registration, such as robustness to diverse noise types, they face limitations when dealing with imbalanced point cloud densities. Both methods rely on correspondence estimation based on local features using DGCNN~\cite{dgcnn}. However, DGCNN's intrinsic lack of surface awareness due to its utilization of the Euclidean metric for k-nearest neighbors introduces challenges~\cite{Sheshappanavar_2020_CVPR_Workshops, Sheshappanavar_maintrack}. To overcome this, authors in~\cite{Sheshappanavar_2020_CVPR_Workshops, Sheshappanavar_maintrack} introduced using ellipsoid query but falls short as it requires extraneous querying. Counter to this we propose Mahalanobis versions of both DCP and DeepUME, introducing the Mahalanobis distance into k-NN to derive surface-aware features that enhance point cloud registration. Furthermore, our proposed approach demonstrates remarkable discriminative capabilities, as evident from its performance in point cloud few-shot classification tasks.

\section{Problem Setting}

We denote \(x_{i}^{T} (i \in [1,M])\) and \(y_{i}^{T} (i \in [1,N])\) as row vectors from matrices \(X \in \mathbb{R}^{M \times 3}\) and \(Y \in \mathbb{R}^{N \times 3}\) respectively. The matrices \(X\) and \(Y\) represent two distinct point clouds, often originating from two scans of the same object (we assume \(Y\) is transformed from \(X\) by an unknown rigid motion). Here, \(x_{i}\) and \(y_{j}\) represent the coordinates of the \(i\)th point in \(X\) and the \(j\)th point in \(Y\), for \(i \in [1,M]\) and \(j \in [1,N]\) respectively. Assuming that there are \(S\) pairs of corresponding points between the two point clouds, the goal of registration is to determine the optimal rigid transformation parameters \(g\) (comprising a rotation matrix \(R \in \mathcal{SO}(3)\) and a translation vector \(t \in \mathbb{R}^{3}\)) that aligns point cloud \(X\) to point cloud \(Y\), as illustrated below:

\begin{equation}
\label{eq:problem}
     \argmin_{R \in \mathcal{SO}(3),\, t \in \mathbb{R}^{3}} \left\Vert d\left(X, g(Y)\right) \right\Vert_{2}^{2}
\end{equation}


The term \(d\big(X,g(Y)\big)\) is a measure of the projection error between \(X\) and the transformed \(Y\), expressed as \(d\big(X,RY + t\big) = \sum_{s=1}^{S} \left\Vert x_{s} -\big(Ry_{s} + t \big) \right\Vert_{2}\), where \(s\) iterates over the \(S\) pairs of corresponding points. The optimization problem in Equation \ref{eq:problem} embodies a well-known ``chicken-and-egg'' scenario: determining the optimal transformation matrix requires knowledge of true correspondences ~\cite{correspondance1,correspondance2}; conversely, accurate correspondences can be established given the optimal transformation matrix. However, the simultaneous resolution of both aspects poses a non-trivial challenge.

\section{Method}


In this paper, we focus on two deep learning-based approaches, specifically DCP ~\cite{DCP} and DeepUME ~\cite{DeepUME}, which address the aforementioned challenge as a global one-shot rigid transformation. This contrasts with the iterative nature of methods like ICP ~\cite{ICP}.  Considering these methods, DGCNN ~\cite{dgcnn} is pivotal for extracting local features crucial for high-quality point matching. Both approaches modify the DGCNN framework to construct a static graph using k-nearest Neighbors (k-NN) for the point cloud. However, this limits the potential for complex semantic space creation. The conventional DGCNN recalculates the k-NN graph after each convolution, projecting features into higher-dimensional space, while the simplified k-NN graph imposition curtails semantic depth. To overcome this limitation, we propose harnessing the \textit{Mahalanobis distance}~\cite{MD} as a statistical lens for enhancing k-NN. This empowers querying based on principal components, emphasizing surface-level k-NN grouping. The Mahalanobis-based k-NN lens extracts features sensitive to surface and corner regions, vital for accurate feature matching and robust transformation estimation. 

\subsection{Mahalanobis k-NN}
\label{sec:mah}

\begin{figure}
    \includegraphics[width=1.0\linewidth]{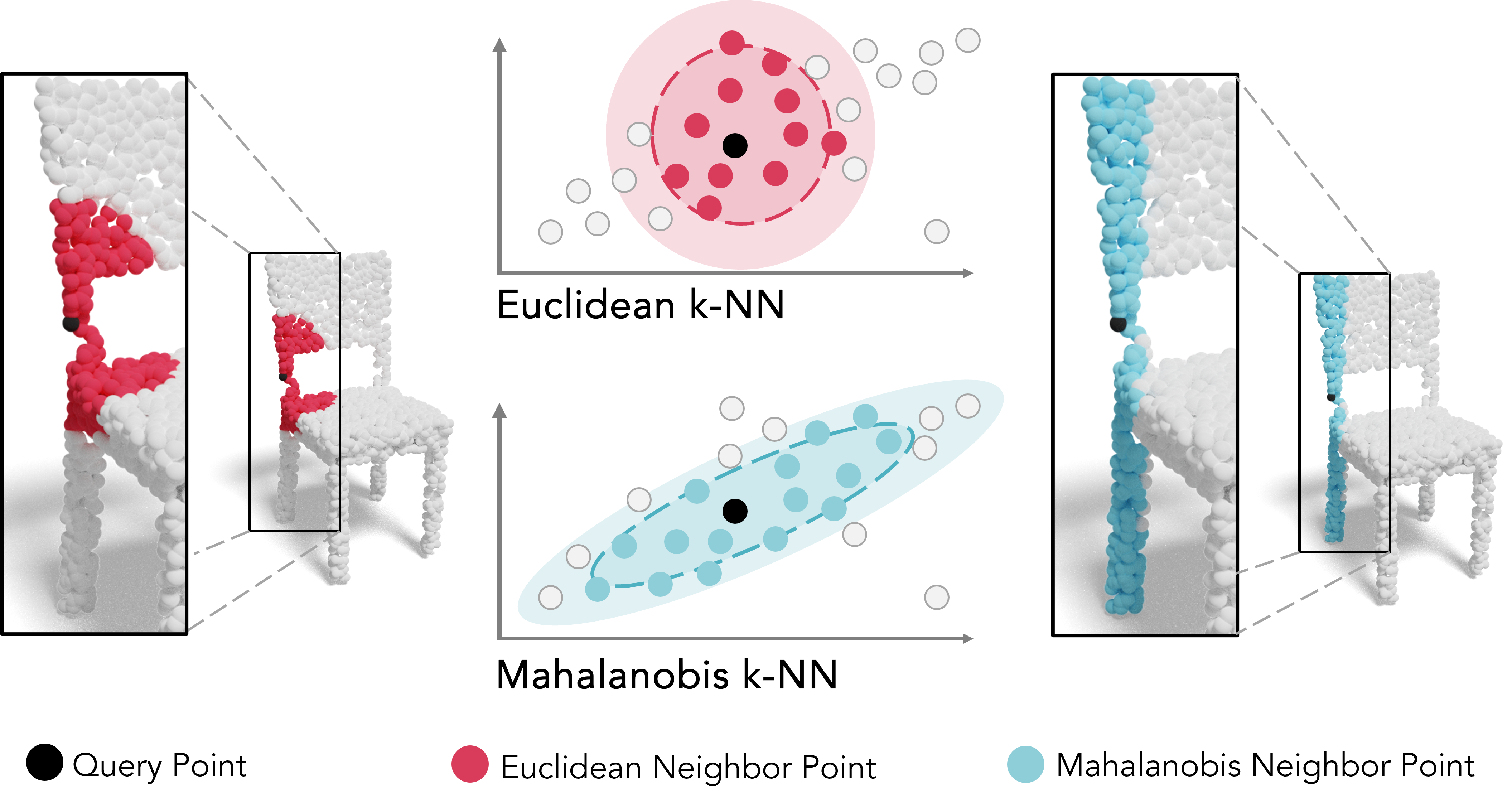}

    \caption{Illustration highlights the distinction between the \textcolor[HTML]{D93C5B}{Red} gradients representing Euclidean distance fields that capture spatial neighbors and the \textcolor[HTML]{8ECED8}{Blue} gradients depicting Mahalanobis distance fields, which consider neighbors concerning the underlying data distribution. To further emphasize the point, we present the influence of both Euclidean and Mahalanobis k-NN on a \textit{chair} point cloud. The \textbf{black} query point is surrounded by Euclidean neighbors shown in \textcolor[HTML]{D93C5B}{red} points and Mahalanobis neighbors in \textcolor[HTML]{8ECED8}{blue} points. The depiction clearly illustrates the impact of Mahalanobis distance, effectively capturing surficial points per the data distribution—vital for precise feature matching.}
    \label{fig:mah-illus}
\end{figure}

In this section, we introduce Mahalanobis distance as a statistical lens for robust feature matching in DCP~\cite{DCP} and DeepUME~\cite{DeepUME}. We modify DGCNN to construct a graph \(\mathcal{G}^{\dagger}\) using Mahalanobis k-nearest neighbors (k-NN) for enhanced feature selection. DGCNN constructs a graph \(\mathcal{G}\), applies nonlinearity for edge values, and performs vertex-wise aggregation (\(\max\) or \(\sum\)) in each layer. Let \(x_{i}^{l}\) be the embedding of point \(i\) in the \(l\)-th layer, and \(h_{\theta}^{l}\) be a nonlinear function parameterized by a shared multi-layer perceptron (MLP). DGCNN can be formulated as:

\begin{equation}
    \label{eq:dgcnn}
    x^{l}_{i} = f \Big( \big\{ h_{\theta}^{l} (x_{i}^{l-1}, x_{j}^{l-1}) \forall j \in \mathcal{N}_{i} \big\}   \Big)
\end{equation}

where \(\mathcal{N}_{i}\) denotes neighbors of vertex \(i\) in \(\mathcal{G}\).

We propose a novel augmentation by integrating Mahalanobis k-nearest neighbors (k-NN) in Equation \ref{eq:dgcnn} to construct graph \(\mathcal{G}^{\dagger}\). This graph prioritizes surface awareness through selection of surficial neighbors \(\mathcal{N}_{i}^{\dagger}\) based on Mahalanobis distance as shown in Figure \ref{fig:mah-illus}.

To compute Mahalanobis distance between points \(x_{i}\) and \(x_{j}\), we use:
\begin{equation}
\label{eq:mah}
    D_M(x_{i}, x_{j}) = \sqrt{(x_{i} - x_{j})^T \cdot C^{-1} \cdot (x_{i} - x_{j})}
\end{equation}
where:
\(D_M(x_{i}, x_{j})\) is Mahalanobis distance between \(x_{i}\) and \(x_{j}\), and \(C\) is covariance matrix of data points. One can argue, geodesic graph-based methods capture similar or sometimes better surficial information. To counter this, we implement vectorized Floyd-Warshall (refer supplementary for implementation detail on vectorized Floyd-Warshall) on the Euclidean-knn graph and observe that the performance is slower and also not as effective as the proposed Mahalanobis variants of DCP and DeepUME. \\

\noindent \textbf{Advantages of Mahalanobis Distance.} Unlike Euclidean distance which treats dimensions independently, Mahalanobis distance \textit{accounts for covariance}, making it suitable for correlated feature datasets. It \textit{scales features invariantly}, accommodating feature scale variations that affect Euclidean distances. This metric excels for datasets with non-spherical or elliptical clusters, considering \textit{data distribution shape}. As data dimensionality increases, Mahalanobis distance is more resilient against the ``curse of dimensionality,'' particularly impactful on Euclidean distances. It demonstrates \textit{robustness to outliers}, focusing on central data distribution. Moreover, Mahalanobis distance enables \textit{customized metric learning}, adapting to dataset peculiarities for improved performance. In machine learning, it serves as a \textit{discriminative metric}, enhancing class separation in classification. Its \textit{statistical interpretation} enhances usability in data analysis. Moreover, it's faster and more robust compared to geodesic-graph-based approaches.

The impact of Mahalanobis distance is visually depicted in Figure \ref{fig:mah-illus}. This surficial awareness proves crucial in feature matching. Incorporating Mahalanobis distance into DCP ~\cite{DCP} and DeepUME ~\cite{DeepUME} yields superior performance in point cloud registration. Discriminative power is affirmed in point-cloud few-shot classification, with pre-trained DGCNN models from vanilla DCP and Mahalanobis DCP showcasing remarkable classification capabilities.

\section{Experiments}
In this section, we showcase the superior performance of our proposed method in the domain of point cloud registration when compared against existing state-of-the-art techniques such as ICP~\cite{ICP}, GO-ICP~\cite{goICP}, FGR~\cite{FGR}, PointNetLK~\cite{pointnetLK}, DCP~\cite{DCP}, and DeepUME~\cite{DeepUME} on publicly available datasets. Our evaluation encompasses a spectrum of benchmarking strategies, encompassing scenarios where one point cloud is notably sparser than the other.
For a fair comparison, we implemented a vectorized Floyd-Warshall algorithm in PyTorch to compute geodesic distances and compare it to our proposed method for surface feature extraction. We further validate the Mahalanobis metric's discriminative power in few-shot learning tasks and showcase robustness through endurance tests. Our model was trained using an Nvidia RTX 3090 GPU and PyTorch 1.11.

\subsection{Point Cloud Registration}
We build upon the model architecture, training strategies, evaluation metrics, and hyperparameters established in DCP~\cite{DCP}~\footnote{\href{https://github.com/WangYueFt/dcp}{https://github.com/WangYueFt/dcp}}. We introduce the Mahalanobis version of DCP (MDCP), detailed in Section~\ref{sec:mah}. MDCP comes in two variants: v1 without a transformer and v2 with a transformer. We further extend our approach to DeepUME~\cite{DeepUME} by proposing the Mahalanobis variant, MDeepUME. For a fair comparison, we benchmark geodesic versions of these models using vectorized Floyd-Warshall on Euclidean k-NN graphs.


%
\noindent \textbf{Dataset.} To facilitate benchmarking, we employ publicly available datasets: ModelNet40~\cite{modelnet}, FAUST~\cite{FAUST}, and Stanford3D (S-3D)\footnote{\href{http:// graphics.stanford.edu/data/3Dscanrep/}{http:// graphics.stanford.edu/data/3Dscanrep/}} . Notably, the latter two datasets are exclusively employed for testing purposes. To ensure methodological consistency, we adhere to the official training/testing splits and settings as stipulated in the original works of DCP~\cite{DCP} and DeepUME~\cite{DeepUME}. Among these datasets, our approach is rigorously evaluated. \\

We evaluate using root mean squared error (RMSE) between ground truth and predicted transformations (lower is better). Angular errors are reported in degrees. These metrics comprehensively assess our approach's performance.


\subsection{Comparison with State-of-the-art Methods}
\begin{table}[!ht]

\caption{Quantitative comparison of state-of-the-art methods benchmarked on ModelNet40~\cite{modelnet} Dataset for three tasks 1) Unseen data, 2) Unseen category, 3) Robustness towards Gaussian Noise; Our proposed MDCP-(v1,v2) and MDeepUME; outperforms its original variants in almost all cases. All evaluation metrics are lower the better. We highlight \textbf{bold} as best and \underline{underline} as second best. \textbf{Note}: $\dagger$ depicts results reproduced on our systems and \textbf{-Geo} refers to Flyod-Warshall geodesic-graph version.}
\label{tab:main}
\resizebox{\linewidth}{!}{%
\begin{tabular}{rcccccc}
\hline\hline
                      & \multicolumn{2}{c}{\textbf{Unseen data}}                                                                  & \multicolumn{2}{c}{\textbf{Unseen Category}}                                                                 & \multicolumn{2}{c}{\textbf{Unseen data + Noise}}                                                                  \\ \cline{2-7} 
                      & \textbf{RMSE(R)}                          & \textbf{RMSE(t)}                          & \textbf{RMSE(R)}                          & \textbf{RMSE(t)}                         & \textbf{RMSE(R)}                          & \textbf{RMSE(t)}                          \\ \hline
\textbf{ICP~\cite{ICP}}          & 29.9148                                 & 0.2909                                  & 29.8764                                & 0.2933                                 & 29.7080                                 & 0.2906                                  \\
\textbf{GO-ICP~\cite{goICP}}       & 11.8523                                 & 0.0257                                  & 13.8657                                & 0.0226                                 & 11.4535                                 & 0.0231                                  \\
\textbf{FGR~\cite{FGR}}          & 9.3628                                  & 0.0139                                  & 9.8490                                 & 0.0135                                & 24.6515                                 & 0.1090                                  \\
\textbf{PointNetLK~\cite{pointnetLK}}   & 15.0954                                 & 0.0221                                  & 17.5021                                 & 0.0280                                 & 16.0049                                  & 0.0216                                  \\ \hline
\textbf{DCP-v1~\cite{DCP}}       & \cellcolor[HTML]{DCECD2}{\underline{2.5457}}    & \cellcolor[HTML]{DCECD2}0.0018         & \cellcolor[HTML]{DCECD2}4.3819          & \cellcolor[HTML]{DCECD2}0.0050          & \cellcolor[HTML]{DCECD2}2.6318          & \cellcolor[HTML]{DCECD2}0.0018          \\
\textbf{DCP-v1{$^\dagger$}~\cite{DCP}}      & \cellcolor[HTML]{DCECD2}2.6581          & \cellcolor[HTML]{DCECD2}0.0009    & \cellcolor[HTML]{DCECD2}6.8386          & \cellcolor[HTML]{DCECD2}0.0035         & \cellcolor[HTML]{DCECD2}2.6048          & \cellcolor[HTML]{DCECD2}{\underline{0.0009}}          \\

\textbf{DCP-v1-Geo}      & \cellcolor[HTML]{DCECD2}2.6001         & \cellcolor[HTML]{DCECD2}{\underline{0.0007}}    & \cellcolor[HTML]{DCECD2}{\underline{3.3326}}          & \cellcolor[HTML]{DCECD2}{\underline{0.0017}}         & \cellcolor[HTML]{DCECD2}{\underline{ 2.1778}}          & \cellcolor[HTML]{DCECD2}0.0019          \\

\textbf{MDCP-v1 (Ours)} & \cellcolor[HTML]{DCECD2}\textbf{1.6614} & \cellcolor[HTML]{DCECD2}\textbf{0.0004}  & \cellcolor[HTML]{DCECD2}\textbf{1.9679} & \cellcolor[HTML]{DCECD2}\textbf{0.0004} & \cellcolor[HTML]{DCECD2}\textbf{1.5889} & \cellcolor[HTML]{DCECD2}\textbf{0.0004} \\ \hline
\textbf{DCP-v2~\cite{DCP}}       & \cellcolor[HTML]{D7E6F4}{\underline{1.1434}}    & \cellcolor[HTML]{D7E6F4}0.0018          & \cellcolor[HTML]{D5E6F5}{\underline{3.1502}}          & \cellcolor[HTML]{D5E6F5}{\underline{0.0050}}         & \cellcolor[HTML]{D7E6F4}{\underline{1.0814}}           & \cellcolor[HTML]{D7E6F4}\textbf{0.0015}   \\
\textbf{DCP-v2{$^\dagger$}~\cite{DCP}}      & \cellcolor[HTML]{D5E6F5}1.4125           & \cellcolor[HTML]{D7E6F4}0.0018    & \cellcolor[HTML]{D5E6F5}4.2246           & \cellcolor[HTML]{D5E6F5}0.0063         & \cellcolor[HTML]{D7E6F4}1.1479          & \cellcolor[HTML]{D7E6F4}{\underline{0.0016}}          \\

\textbf{DCP-v2-Geo}      & \cellcolor[HTML]{D7E6F4}1.4478       & \cellcolor[HTML]{D7E6F4}{\underline{0.0017}}    & \cellcolor[HTML]{D7E6F4}5.5233          & \cellcolor[HTML]{D7E6F4}0.0051         & \cellcolor[HTML]{D7E6F4}2.1668          & \cellcolor[HTML]{D7E6F4}0.0019          \\

\textbf{MDCP-v2 (Ours)} & \cellcolor[HTML]{D7E6F4}\textbf{0.9468} & \cellcolor[HTML]{D7E6F4}\textbf{0.0004} & \cellcolor[HTML]{D5E6F5}\textbf{1.3849} & \cellcolor[HTML]{D5E6F5}\textbf{0.0049} & \cellcolor[HTML]{D7E6F4}\textbf{0.8481} & \cellcolor[HTML]{D7E6F4}0.0043           \\ \hline
\textbf{DeepUME{$^\dagger$}~\cite{DeepUME}}      & \cellcolor[HTML]{FFFFC7}0.0062         & \cellcolor[HTML]{FFFFC7}\textbf{0}                 & \cellcolor[HTML]{FFFFC7}0.0183               & \cellcolor[HTML]{FFFFC7}0.00019               & \cellcolor[HTML]{FFFFC7}2.5263 & \cellcolor[HTML]{FFFFC7}\textbf{0.0006} \\

\textbf{DeepUME-Geo}    & \cellcolor[HTML]{FFFFC7}{\underline{0.0051}}  & \cellcolor[HTML]{FFFFC7}\textbf{0}                 & \cellcolor[HTML]{FFFFC7}{\underline{0.0106}} & \cellcolor[HTML]{FFFFC7}{\underline{0.00014}}             & \cellcolor[HTML]{FFFFC7}{\underline{2.0086}}       & \cellcolor[HTML]{FFFFC7}{\underline{0.0008}} \\ 

\textbf{MDeepUME}    & \cellcolor[HTML]{FFFFC7}\textbf{0.0023}  & \cellcolor[HTML]{FFFFC7}\textbf{0}                 & \cellcolor[HTML]{FFFFC7} \textbf{0.0098} & \cellcolor[HTML]{FFFFC7}    \textbf{0.00009}             & \cellcolor[HTML]{FFFFC7}\textbf{1.9946}          & \cellcolor[HTML]{FFFFC7}\textbf{0.0006} \\ \hline\hline
\end{tabular}%
}
\end{table}


\noindent{\textbf{Results on ModelNet40 Dataset~\cite{modelnet}.}} Comprehensive results are presented in Table \ref{tab:main}, offering a thorough comparison between our method, state-of-the-art alternatives and their geodesic versions. The presented outcomes highlight the superior performance of our method across all evaluation metrics and in diverse settings. Our benchmarking encompasses scenarios including \textit{Unseen-data} evaluation, where the train-test split outlined in ModelNet40~\cite{modelnet} is utilized; \textit{Unseen category} evaluation, where the first 20 classes of ModelNet40 are employed for training and the remaining for testing; and a robustness assessment against \textit{Gaussian Noise} $\mathcal{N}(0, 0.01)$, which is clipped to the interval [-$ 0.05, 0.05 $] before being added to the point cloud. Notably, a point density of 1024 is consistently sampled in all aforementioned settings. \textbf{Note:} Our assessment of state-of-the-art methods is derived from~\cite{DCP}, as our efforts to reproduce the open-source code yielded highly comparable results with minimal standard deviation.

In each case, a rigid transformation is introduced along each axis, with rotation uniformly sampled in the range [$ 0^{\circ},45^{\circ} $] and translation within [-$ 0.5, 0.5 $]. The input to the network consists of both the original point cloud and the transformed point cloud through the rigid motion. This input is subsequently evaluated against the known ground truth in both our proposed MDCP and MDeepUME methods.

\begin{figure*}[!ht]
    \center
    \includegraphics[width=\linewidth]{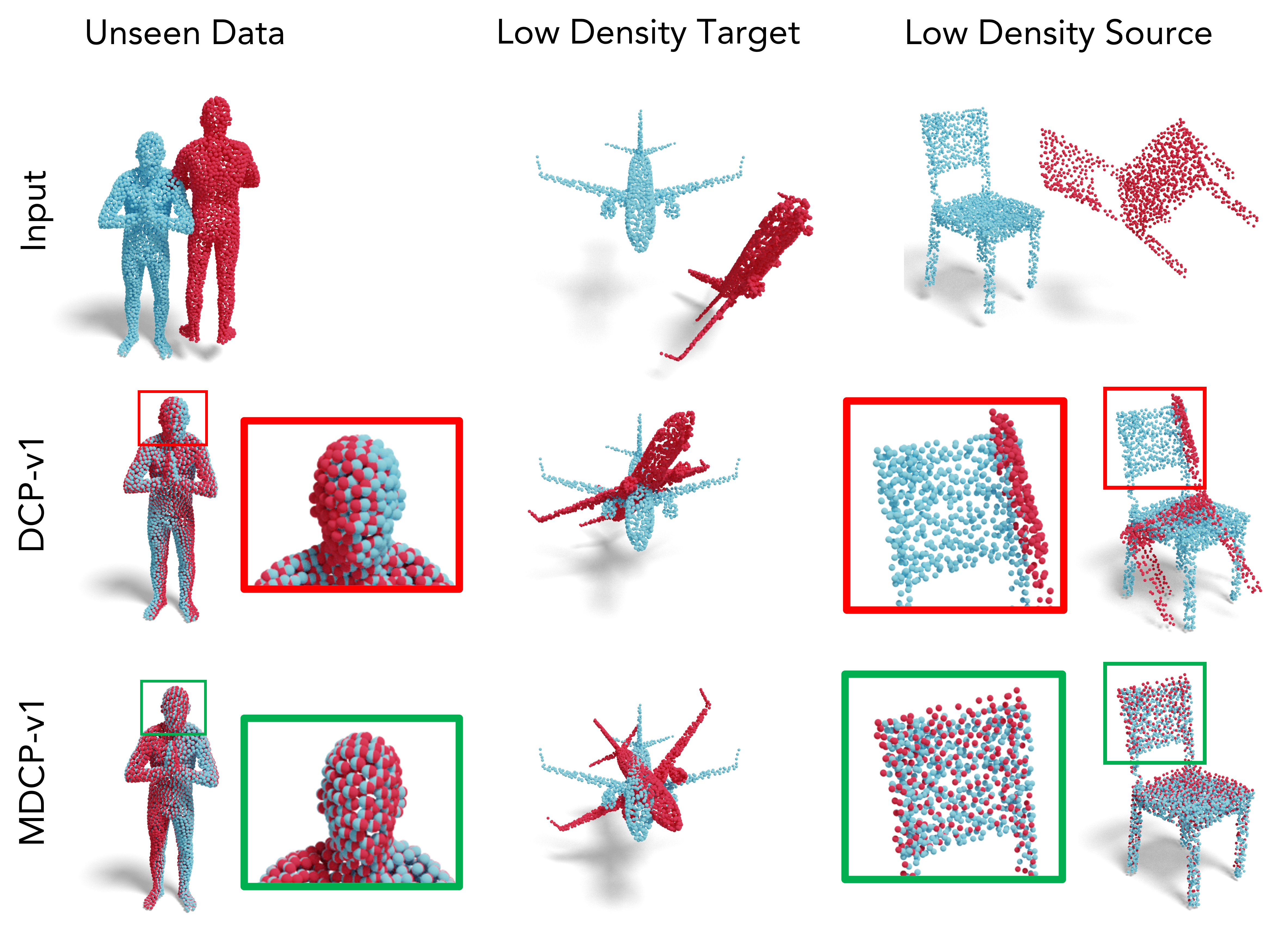}

    \caption{Evaluation of Robustness for varied test cases. Both DCP-v1~\cite{DCP} and proposed MDCP-v1 trained on ModelNet40~\cite{modelnet} dataset. Our evaluation encompasses three distinct scenarios: 1) Evaluation on unseen data, specifically a man holding palms together from the FAUST~\cite{FAUST} dataset; 2) Assessment of low-density source point clouds, exemplified by an airplane; and 3) Examination of low-density target point clouds, illustrated by a chair. In our visual analysis, regions highlighted in \textcolor{red}{red} highlights the vulnerabilities observed in vanilla DCP-v1~\cite{DCP}, while regions in \textcolor{green}{green} accentuate the pronounced efficacy demonstrated by the proposed MDCP-v1 over the conventional DCP-v1.}
    \label{fig:reg}
\end{figure*}

\begin{figure}[!ht]
    \center
    \includegraphics[width=1\linewidth]{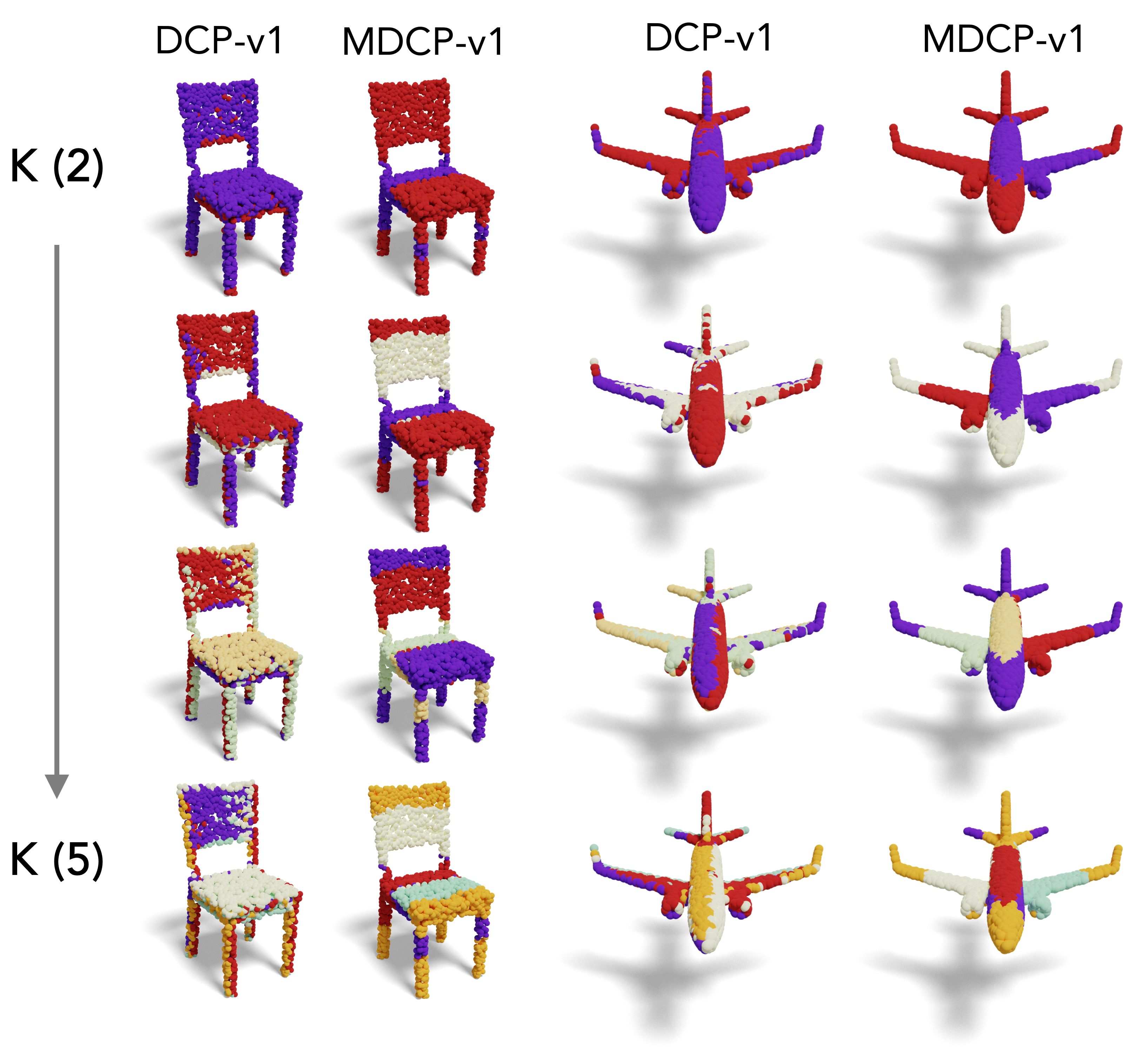}
    \caption{We present a compelling demonstrations showcasing the surface awareness within the proposed MDCP-v1, in contrast to the Euclidean approach in DCP-v1\cite{DCP}. Each row within the illustration corresponds to K values ranging from 2 to 5, effectively exemplifying the pronounced surficial awareness embodied by MDCP-v1. This surficial awareness is pivotal for robust point cloud registration tasks.}
    \label{fig:seg}
\end{figure}

Furthermore, to substantiate the assertion that Mahalanobis distance operates as a statistical lens for enhanced feature matching, we undertake KMeans~\cite{kmeans} clustering within the initial embedding space of DGCNN. This embedding space, a 64-dimensional domain following the first edge-convolution~\cite{dgcnn}, is employed both in DCP-v1 and MDCP-v1. Our findings are presented visually through Figure \ref{fig:seg}. The depicted results unequivocally showcase the preservation of surface awareness in the case of the Mahalanobis variant, even in scenarios involving overclustering (e.g., K=5). This outcome stands in contrast to its Euclidean counterpart. Particularly is the consistent maintenance of surface awareness throughout various clustering cases, particularly evident in the context of the Airplane dataset. This affirmation highlights that the Mahalanobis version of DCP learns to discern and emphasize surface-aware regions that are crucial in achieving precise feature matching.

\begin{table}[!ht]
\centering

\caption{We present the outcomes of few-shot classification experiments conducted on the ModelNet40~\cite{modelnet} dataset, as well as on three distinct splits (OBJ ONLY, OBJ+BG, and PB75) of the ScanObjectNN~\cite{SONN} dataset. Our proposed DCGNN employed in MDCP-(v1,v2) consistently surpasses its original variants by a substantial margin across all few-shot settings and datasets. The reported quantitative results in terms of accuracies (\%) are presented below, with values in \textbf{bold} indicating the best performance. Additionally, we provide the standard deviation values derived from 50 runs for a comprehensive understanding of the results.}
\label{tab:fsl}

\resizebox{1.0\linewidth}{!}{%
\begin{tabular}{rcccccccc}

\hline\hline
\multicolumn{9}{c}{\cellcolor[HTML]{EFEFEF}\textbf{ModelNet40~\cite{modelnet}}}                                                                                                       \\ \hline
                     & \multicolumn{4}{c}{\textbf{5 way}}                                          & \multicolumn{4}{c}{\textbf{10 way}}                                         \\ \cline{2-9} 
\textbf{Method}      & \multicolumn{2}{c}{\textbf{10 shot}} & \multicolumn{2}{c}{\textbf{20 shot}} & \multicolumn{2}{c}{\textbf{10 shot}} & \multicolumn{2}{c}{\textbf{20 shot}} \\ \hline
\textbf{DCP-v1~\cite{DCP}}      & 67.6              & $\pm$6.91             & 69                & $\pm$7.87             & 50.5              & $\pm$7.74             & 57.1              & $\pm$5.01             \\
\textbf{MDCP-v1 (Ours)} & \textbf{75.2}     & $\pm$\textbf{7.3}     & \textbf{80}       & $\pm$\textbf{7.74}    & \textbf{71.0}       & $\pm$\textbf{4.16}    & \textbf{70.9}     & $\pm$\textbf{5.72}    \\
\cellcolor[HTML]{E3FDE3}\textbf{Increment in \% }   &\cellcolor[HTML]{E3FDE3} \textbf{11.24}    &\cellcolor[HTML]{E3FDE3}                  & \cellcolor[HTML]{E3FDE3}\textbf{15.94}    & \cellcolor[HTML]{E3FDE3}                 &\cellcolor[HTML]{E3FDE3} \textbf{40.59}    & \cellcolor[HTML]{E3FDE3}                 &\cellcolor[HTML]{E3FDE3} \textbf{24.17}    & \cellcolor[HTML]{E3FDE3}                 \\ \hline
\textbf{DCP-v2~\cite{DCP}}      & 76.8              &$\pm$4.01             & 79.8              & $\pm$6.17             & 60.8              & $\pm$4.93             & 69.15             & $\pm$5.7              \\
\textbf{MDCP-v2 (Ours)} & \textbf{82.5}     & $\pm$\textbf{6.96}    & \textbf{81.8}     & $\pm$\textbf{7.78}    & \textbf{67.9}     & $\pm$\textbf{5.43}    & \textbf{72.25}    & $\pm$\textbf{6.33}    \\
\cellcolor[HTML]{E3FDE3}\textbf{Increment in \% }   & \cellcolor[HTML]{E3FDE3}\textbf{7.42}     &\cellcolor[HTML]{E3FDE3}                  & \cellcolor[HTML]{E3FDE3}\textbf{2.51}     &\cellcolor[HTML]{E3FDE3}                  &\cellcolor[HTML]{E3FDE3} \cellcolor[HTML]{E3FDE3}\textbf{11.68}    & \cellcolor[HTML]{E3FDE3}                 & \cellcolor[HTML]{E3FDE3}\textbf{4.48}     &\cellcolor[HTML]{E3FDE3}\\ \hline
\multicolumn{9}{c}{\cellcolor[HTML]{EFEFEF}\textbf{ScanObjectNN OBJ ONLY~\cite{SONN}}}                                                                                                       \\ \hline
\textbf{DCP-v1~\cite{DCP}}      & 51.7              & $\pm$8.08             & 54.8              & $\pm$8.82             & 32.15             & $\pm$3.56             & 34.55             & $\pm$3.17             \\
\textbf{MDCP-v1 (Ours)} & \textbf{54.7}     & $\pm$\textbf{7.47}    & \textbf{57.1}     & $\pm$\textbf{6.51}    & \textbf{39.0}       & $\pm$\textbf{4.88}    & \textbf{41.2}     & $\pm$\textbf{5.72}    \\
\cellcolor[HTML]{E3FDE3}\textbf{Increment in \% }   &\cellcolor[HTML]{E3FDE3} \textbf{5.80}     & \cellcolor[HTML]{E3FDE3}                 & \cellcolor[HTML]{E3FDE3}\textbf{4.20}     & \cellcolor[HTML]{E3FDE3}                 &\cellcolor[HTML]{E3FDE3} \textbf{21.31}    &\cellcolor[HTML]{E3FDE3}                  & \cellcolor[HTML]{E3FDE3}\textbf{19.25}    & \cellcolor[HTML]{E3FDE3}                 \\ \hline
\textbf{DCP-v2~\cite{DCP}}      & 55.1              & $\pm$9.57             & 56.2              & $\pm$6.8              & 38.45             & $\pm$5.08             & 42.05             & $\pm$4.93             \\
\textbf{MDCP-v2 (Ours)} & \textbf{55.4}     & $\pm$\textbf{12.11}   & \textbf{62.4}     & $\pm$\textbf{7.91}    & \textbf{42.85}    & $\pm$\textbf{3.37}    & \textbf{46.15}    & $\pm$\textbf{2.96}    \\
\cellcolor[HTML]{E3FDE3} \textbf{Increment in \% }   & \cellcolor[HTML]{E3FDE3} \textbf{0.54}     & \cellcolor[HTML]{E3FDE3}                 & \cellcolor[HTML]{E3FDE3} \textbf{11.03}    &  \cellcolor[HTML]{E3FDE3}                & \cellcolor[HTML]{E3FDE3} \textbf{11.44}    &   \cellcolor[HTML]{E3FDE3}               & \cellcolor[HTML]{E3FDE3} \textbf{9.75}     & \cellcolor[HTML]{E3FDE3}                 \\ \hline
\multicolumn{9}{c}{\cellcolor[HTML]{EFEFEF}\textbf{ScanObjectNN OBJ + BG~\cite{SONN}}}                                                                                                       \\ \hline
\textbf{DCP-v1~\cite{DCP}}      & 50.3              & $\pm$5.25             & 48.2              & $\pm$7.35             & 29.45             & $\pm$5.34             & 32.15             & $\pm$2.9              \\
\textbf{MDCP-v1 (Ours)} & \textbf{62.2}     & $\pm$\textbf{6.73}    & \textbf{60.2}     & $\pm$\textbf{7.54}    & \textbf{37.25}    & $\pm$\textbf{3.26}    & \textbf{41.45}    & $\pm$\textbf{3.21}    \\
\cellcolor[HTML]{E3FDE3}\textbf{Increment in \% }   &\cellcolor[HTML]{E3FDE3} \textbf{23.66}    &\cellcolor[HTML]{E3FDE3}                  & \cellcolor[HTML]{E3FDE3}\textbf{24.90}    & \cellcolor[HTML]{E3FDE3}                 & \cellcolor[HTML]{E3FDE3}\textbf{26.49}    & \cellcolor[HTML]{E3FDE3}                 & \cellcolor[HTML]{E3FDE3}\textbf{28.93}    & \cellcolor[HTML]{E3FDE3}                 \\ \hline
\textbf{DCP-v2~\cite{DCP}}      & 52.3              & $\pm$6.92             & 55                & $\pm$8.16             & 33.8              & $\pm$4.31             & 36.25             & $\pm$4.57             \\
\textbf{MDCP-v2 (Ours)} & \textbf{58.8}     & $\pm$\textbf{8.07}    & \textbf{62.8}     & $\pm$\textbf{6.22}    & \textbf{41.9}     & $\pm$\textbf{6.2}     & \textbf{44.6}     & $\pm$\textbf{4.1}     \\
\cellcolor[HTML]{E3FDE3}\textbf{Increment in \% }   & \cellcolor[HTML]{E3FDE3}\textbf{12.43}    & \cellcolor[HTML]{E3FDE3}                 & \cellcolor[HTML]{E3FDE3}\textbf{14.18}    & \cellcolor[HTML]{E3FDE3}                 & \cellcolor[HTML]{E3FDE3}\textbf{23.96}    &\cellcolor[HTML]{E3FDE3}                  &\cellcolor[HTML]{E3FDE3} \textbf{23.03}    & \cellcolor[HTML]{E3FDE3}                 \\ \hline
\multicolumn{9}{c}{\cellcolor[HTML]{EFEFEF}\textbf{ScanObjectNN PB75~\cite{SONN}}}                                                                                                           \\ \hline
\textbf{DCP-v1~\cite{DCP}}      & 46.1              & $\pm$8.31             & 45.2              & $\pm$9.37             & 25.2              & $\pm$3.78             & 28.45             & $\pm$4.5              \\
\textbf{MDCP-v1 (Ours)} & \textbf{54.8}     & $\pm$\textbf{7.35}    & \textbf{51.5}     & $\pm$\textbf{7.08}    & \textbf{34.05}    & $\pm$\textbf{4.2}     & \textbf{37.65}    & $\pm$\textbf{4.69}    \\
\cellcolor[HTML]{E3FDE3}\textbf{Increment in \% }   & \cellcolor[HTML]{E3FDE3}\textbf{18.87}    & \cellcolor[HTML]{E3FDE3}                 & \cellcolor[HTML]{E3FDE3}\textbf{13.94}    & \cellcolor[HTML]{E3FDE3}                 & \cellcolor[HTML]{E3FDE3}\textbf{35.12}    & \cellcolor[HTML]{E3FDE3}                 & \cellcolor[HTML]{E3FDE3}\textbf{32.34}    & \cellcolor[HTML]{E3FDE3}                 \\ \hline
\textbf{DCP-v2~\cite{DCP}}      & 49.6              & $\pm$6.64             & 49.6              & $\pm$8.7              & 27.85             & $\pm$3.75             & 32.35             & $\pm$4.03             \\
\textbf{MDCP-v2 (Ours)} & \textbf{54.7}     & $\pm$\textbf{5.89}    & \textbf{53.2}     & $\pm$\textbf{9.46}    & \textbf{34.45}    & $\pm$\textbf{3.48}    & \textbf{41.15}    & $\pm$\textbf{4.87}    \\
\cellcolor[HTML]{E3FDE3}\textbf{Increment in \% }   &\cellcolor[HTML]{E3FDE3} \textbf{10.28}    & \cellcolor[HTML]{E3FDE3}                 & \cellcolor[HTML]{E3FDE3}\textbf{7.26}     & \cellcolor[HTML]{E3FDE3}                 & \cellcolor[HTML]{E3FDE3}\textbf{23.70}    & \cellcolor[HTML]{E3FDE3}                 & \cellcolor[HTML]{E3FDE3}\textbf{27.20}    & \cellcolor[HTML]{E3FDE3}                 \\ \hline \hline
\end{tabular}%
}
\end{table}

\noindent{\textbf{Few-shot Evaluation.}} The discriminative prowess of the proposed Mahalanobis distance is evaluated through Table~\ref{tab:fsl}, illustrating the performance of features learned by both DCP-(v1,v2)~\cite{DCP} and our proposed MDCP-(v1,v2) in the context of few-shot evaluation tasks. The evaluation adheres to a \(k\)-way, \(m\)-shot few-shot strategy detailed in ~\cite{CrossPoint}. The models are initially trained for the point-cloud registration task on the ModelNet40 dataset, and subsequently assessed for few-shot classification tasks on the ModelNet40 test set. Furthermore, evaluations extend to three distinct splits, namely OBJ ONLY, OBJ + BG, and PB75, from the ScanObjectNN~\cite{SONN} dataset.

A striking observation is the substantial performance enhancement of the Mahalanobis version of DCP, which surpasses its original variant by a significant margin across all few-shot tasks. Summarizing the performance of Mahalanobis versions (v1,v2) as reported in Table \ref{tab:fsl}, we increment across all few-shot tasks on the ModelNet40~\cite{modelnet} dataset of \textit{(22.98, 6.52)}. In the context of the ONLY OBJ split, an average increment of \textit{(12.68, 8.19)} is recorded, while for the OBJ + BG split, the increments are \textit{(25.99, 18.40)}. Similarly, for the PB75 split, increments of \textit{(25.06, 17.11)} are documented. \textbf{Note}: (a, b) denotes average accuracy increment in \% for MDCP-(v1,v2).

From the results, two insights emerge: Firstly, v1 without a transformer outperforms v2 with a transformer, suggesting that transformers may not generalize as effectively with limited data. Secondly, its worth noting that even when evaluated on unseen ScanObjectNN~\cite{SONN} data, which was not included during the training phase, a consistent increment is observed across all splits.

\subsection{Endurance Test}

\begin{table}[!ht]
\centering

\caption{Quantitative comparison to evaluate the robustness of the proposed MDCP-(v1,v2) against its original variants, benchmarked on the ModelNet40~\cite{modelnet} dataset. This comparison is conducted across various settings for the point-cloud registration task, particularly when the target and source point cloud exhibit low point density. Throughout the comparison, our approach consistently demonstrates evident superiority in performance. The \textbf{bold} typeface denotes the best results. It is important to note that all evaluation metrics follow a ``lower is better'' criterion, and the \textbf{RMSE(t)} metric is reported in units of \(10^{-2}\). \textbf{Note:} \textbf{-Geo} refers to Flyod-Warshall geodesic-graph version.}
\label{tab:SL-TL}
\resizebox{1.0\linewidth}{!}{%
\begin{tabular}{crcccccc}
\hline\hline
                                                            \multicolumn{8}{c}{\cellcolor[HTML]{EFEFEF}\textbf{Target Low Density ModelNet40~\cite{modelnet}}}           \\ \hline
                                                            & \multicolumn{1}{l}{} & \multicolumn{2}{c}{\textbf{Unseen Data}}          & \multicolumn{2}{c}{\textbf{Unseen Category}}     & \multicolumn{2}{c}{\textbf{Noise}}
                                                            \\ \cline{3-8} 
                                                            & \multicolumn{1}{l}{} & \multicolumn{1}{c}{\textbf{RMSE(R)}} & \multicolumn{1}{c}{\textbf{RMSE(t)}} & \multicolumn{1}{c}{\textbf{RMSE(R)}} & \multicolumn{1}{c}{\textbf{RMSE(t)}} & \multicolumn{1}{c}{\textbf{RMSE(R)}} & \multicolumn{1}{c}{\textbf{RMSE(t)}} \\ \hline
                                                            
\multirow{6}{*}{\rotatebox[origin=c]{90}{\textbf{1024}}}    & \textbf{DCP-v1~\cite{DCP}}      & 42.0026                             & 10.865                             & 8.3181                            & 1.2206                             & 32.0181                              & 0.9616                             \\

    & \textbf{DCP-v1-Geo}      & 38.1172                             & 10.863                             & 6.0088                            & 1.2806                             & 39.9996                              & \textbf{0.9801}                             \\
   & \textbf{MDCP-v1}     & \textbf{32.6880}                    & \textbf{10.977}                    & \textbf{5.8425}                    & \textbf{1.2096}                    & \textbf{24.9222}                    & 1.0062                    \\
                                                            & \textbf{DCP-v2~\cite{DCP}}      & 68.0209      & 3.6777  & 14.9977    & \textbf{2.7765} & 66.9599  & 3.0532  \\

    & \textbf{DCP-v2-Geo}      & 44.6676                             & 3.7786                             & 14.7762                            & 8.1678                             & 48.2278                              & 2.9987                            \\
    
                                                            & \textbf{MDCP-v2}     & \textbf{34.9602}      & \textbf{3.3871}  & \textbf{14.3836} & 6.1023      & \textbf{25.8595}  & \textbf{2.8308}   \\ \hline

\multirow{6}{*}{\rotatebox[origin=c]{90}{\textbf{512}}}    & \textbf{DCP-v1~\cite{DCP}}      & 54.5579                             & 1.6058                             & \textbf{8.31497}                     & 1.6812                             & 50.4602                             & 1.4233                             \\

    & \textbf{DCP-v1-Geo}      & 48.1628                             & 1.7898                             & 8.1065                            & 1.6711                             & 49.7866                              & 1.5661                             \\
    
                                                            & \textbf{MDCP-v1}     & \textbf{42.3370}      & \textbf{1.4918} & 8.4701  & \textbf{1.6028}  & \textbf{38.0739} & \textbf{1.3168}  \\
                                                     & \textbf{DCP-v2~\cite{DCP}}      & 80.8099   & 8.9802    & \textbf{23.8370}      & \textbf{3.7857}  & 77.4393   & 8.7660    \\

    & \textbf{DCP-v2-Geo}      & 77.8756                             & 6.7752                             & 21.8861                            & 3.6544                             & 48.2165                              & 8.7765                            \\
       & \textbf{MDCP-v2}     & \textbf{47.3271}                    & \textbf{3.9172}                    & 28.7305                             & 8.4551                             & \textbf{44.3240}                    & \textbf{3.4637}                    \\ \hline

                               \multicolumn{8}{c}{\cellcolor[HTML]{EFEFEF}\textbf{Source Low Density ModelNet40~\cite{modelnet}}}                                        \\ \hline
\multirow{6}{*}{\rotatebox[origin=c]{90}{\textbf{1024}}} & \textbf{DCP-v1~\cite{DCP}}      & 43.2637                            & 1.1146                             & 8.304669                    & 1.228                              & 38.6104                            & 1.0306                             \\

    & \textbf{DCP-v1-Geo}      & 36.6378                             & 1.1123                             & \textbf{8.0011}                           & 1.2761                             & 39.1654                             & 1.1756                             \\
    
                               & \textbf{MDCP-v1}                               & \textbf{31.3686}                   & \textbf{1.1035}                    & 8.3409                               & \textbf{1.2247}                    & \textbf{26.9862}                   & \textbf{1.0158}                    \\
                               & \textbf{DCP-v2~\cite{DCP}}                                & 65.4777                            & 3.1448                             & 16.1993                              & \textbf{1.9267}                    & 61.1659                            & 2.6327                             \\

    & \textbf{DCP-v2-Geo}      & 57.1998                             & 3.0098                             & 14.7176                            & 6.7861                             & 49.1109                              & 2.9987                            \\
                               & \textbf{MDCP-v2}                               & \textbf{38.6540}                   & \textbf{2.8759}                    & \textbf{7.4574}                      & 5.6750                              & \textbf{27.1185}                   & \textbf{2.6114}                    \\ \hline 
      
\multirow{6}{*}{\rotatebox[origin=c]{90}{\textbf{512}}} & \textbf{DCP-v1~\cite{DCP}}      & 55.2426                            & 1.5553                             & 8.4311                               & 1.4074                             & 51.7135                            & 1.3908                             \\

    & \textbf{DCP-v1-Geo}      & \textbf{38.1172}                             & 1.0863                             & \textbf{6.0088}                            & 1.2806                             & 39.9996                              & \textbf{0.9801}                             \\
                               & \textbf{MDCP-v1}                               & 42.4809                   & \textbf{1.4700}                    & 7.8250                      & 1.1089                             & \textbf{39.6634}                   & \textbf{1.3272}                    \\
                               & \textbf{DCP-v2~\cite{DCP}}                                & 71.8377                            & 7.2431                             & \textbf{22.3028}                     & 4.8566                             & 71.1976                            & 6.9627                             \\
    & \textbf{DCP-v2-Geo}      & \textbf{44.6676}                             & 3.7698                             & 26.8861                           & 4.9908                             & 65.9981                              & 5.0098                            \\
                               & \textbf{MDCP-v2}                               & 50.1560                    & \textbf{3.6758}                   & 29.2851                              & \textbf{4.4566}                    & \textbf{45.7104}                   & \textbf{3.3395}                    \\ \hline\hline
\end{tabular}%
}
\end{table}

\noindent{\textbf{Efficiency towards Point-density.}} Table~\ref{tab:SL-TL} evaluates MDCP's robustness against varying point densities, simulating real-world sensor data~\cite{survey}. One point cloud (source or target) is downsampled by half (e.g., 2048 → 1024). Evaluation metrics use the same number of points for both clouds (e.g., source with 1024 points and target with 2048 points are used to compute transformation, then applied to original points for error calculation).

Figure~\ref{fig:reg} compares DCP-v1~\cite{DCP} and MDCP-v1. When densities are similar (man's face), both perform well. However, MDCP-v1 demonstrates superior robustness under varying densities (chair). This highlights its effectiveness in real-world applications where source or target point clouds may have different densities due to factors like sensor capabilities or data acquisition methods (e.g., SFM-LiDAR fusion).

\textbf{Note:} Source point cloud is a transformed version of $g(Y)$, and the goal is to estimate $g$ for perfect alignment with target $X$.

\noindent{\textbf{Efficiency towards Point-density.}} Table \ref{tab:SL-TL} illustrates the robustness of the proposed Mahalanobis version of DCP~\cite{DCP} in the context of point-cloud registration tasks that simulate real-world scenarios. These scenarios involve two point clouds originating from distinct sensors~\cite{survey}, potentially exhibiting differing point densities. In this experimental setup, one of the two point clouds, either the source or target, is downsampled by half relative to the other. As depicted in Table \ref{tab:SL-TL}, the value 1024 signifies that either the target or source point cloud is downsampled to half its original size (e.g., 2048 \(\to\) 1024). \textbf{Note:} To compute evaluation metrics, both source and target point clouds are ensured to have the same number of points. Thus, rigid-transformation parameters are computed using the downsampled point cloud and applied to the original-scale points (e.g., \(X_{2048}\) and \(Y_{1024}\) are used to generate transformation parameters \(g\), which are then utilized to evaluate RMSE(\(X_{2048}\), \(g(Y_{2048})\))).

This scenario is effectively illustrated in Figure \ref{fig:reg}, where the performance comparison between DCP-v1~\cite{DCP} and MDCP-v1 is particularly insightful. In cases where the point clouds have same density, such as in the highlighted regions depicting a man's face, DCP-v1 and MDCP-v1 exhibit relatively close performance. However, the subtleties become more pronounced when subjected to the aforementioned settings. Specifically, DCP-v1 struggles to accurately estimate rigid transformations, leading to subtle angular discrepancies as evident from the highlighted regions. In contrast, the proposed MDCP-v1 showcases a robustness to varying point densities, as demonstrated through the highlighted regions portraying a chair.

\textbf{Note:} The source point cloud is a transformed version of \(g(Y)\), wherein the objective is to estimate the rigid transformation parameters \(g\) that align it perfectly with the target point cloud \(X\).


\begin{table}[!ht]
\centering
\caption{Quantitative comparison to evaluate the robustness of the proposed MDeepUME against its original variant, benchmarked on the ModelNet40~\cite{modelnet}, FAUST~\cite{FAUST} and Stanford 3D repository dataset. This comparison is conducted across various noise types proposed in DeepUME~\cite{DeepUME} for the point-cloud registration task. Throughout the comparison, our approach consistently demonstrates evident superiority in performance. The \textbf{bold} typeface is utilized to denote the best results. It is important to note that all evaluation metrics follow a ``lower is better'' criterion, and the \textbf{RMSE(t)}, chamfer-distance~\cite{ChamferDistance} (\textbf{CD}), and hausdorff-distance~\cite{HausdorffDistance} (\textbf{HD}) metric are reported in units of \(10^{-2}\).}

\label{tab:noise}
\resizebox{1.0\linewidth}{!}{%
\begin{tabular}{cccccc|cccc}
\hline\hline
                                                   &    \textbf{Noise-types}      & \multicolumn{4}{c}{\textbf{DeepUME~\cite{DeepUME}}}    & \multicolumn{4}{c}{\textbf{MDeepUME}} \\ \cline{3-10} 
                                                   &      & \textbf{CD}       & \textbf{HD}       & \textbf{RMSE(R)}  & \textbf{RMSE(t)}  & \textbf{CD}       & \textbf{HD}       & \textbf{RMSE(R)}  & \textbf{RMSE(t)} \\ \hline
\multirow{4}{*}{\rotatebox[origin=c]{90}{\textbf{MN 40}}}& \textbf{Bernoulli}         & 1.113           & 8.6300            & 46.39461          & \textbf{1.5381} & \textbf{1.1120}  & \textbf{8.6200}   & \textbf{45.7815} & 1.5398         \\
                                               & \textbf{Gaussian}          & 0.1913          & 1.2445          & 2.526296          & \textbf{0.0625} & \textbf{0.1900}   & \textbf{1.2400}   & \textbf{1.9946} & 0.0642         \\
                                                      & \textbf{Sampling}    & 0.6317          & 5.6678          & 31.39413          & \textbf{0.9197} & \textbf{0.6290}  & \textbf{5.6620}  & \textbf{30.7709} & 0.9236     \\
                                                     & \textbf{Z-Intersection} & 1.2237      & 11.3676      & 89.48223     & \textbf{0.9003} & \textbf{1.2230}  & \textbf{11.3670}  & \textbf{87.4486} & 0.911     \\ \hline
\multirow{4}{*}{\rotatebox[origin=c]{90}{\textbf{FAUST}}}                                                & \textbf{Bernoulli}         & \textbf{0.2608} & \textbf{2.7165} & 11.30384          & \textbf{2.2294} & 0.2728          & 2.7818          & \textbf{10.3421} & 2.1964         \\
                                                          & \textbf{Gaussian}          & 0.1243          & 0.9342          & 1.819818          & \textbf{0.1085} & \textbf{0.1230}  & \textbf{0.9300}   & \textbf{1.7029}  & 0.2692         \\
                                                                                                         & \textbf{Sampling}          & 0.1026          & 0.9342          & 5.6842            & \textbf{1.172}  & \textbf{0.0941} & \textbf{0.9320}  & \textbf{3.7985} & 0.9461         \\
                                                                                                         & \textbf{Z-Intersection} & 0.2064          & 2.3368          & 12.3721           & \textbf{2.2183} & \textbf{0.2000}    & \textbf{2.3350}  & \textbf{11.9663} & 2.1383         \\  \hline
\multirow{4}{*}{\rotatebox[origin=c]{90}{\textbf{S-3D}}}& \textbf{Bernoulli}         & \textbf{0.2345} & 10.4470           & \textbf{5.951346} & \textbf{1.2268} & 0.2470           & \textbf{10.1853} & 6.4044          & 1.2266         \\
                                                                                                         & \textbf{Gaussian}          & 0.1068          & 0.9477          & 0.392542          & \textbf{0.0591} & \textbf{0.1040}  & \textbf{0.9410}  & \textbf{0.3106} & 0.0604         \\ 
                                                                                                         & \textbf{Sampling}          & 0.0908          & 8.5766          & 6.709815          & \textbf{0.7145} & \textbf{0.0900}   & \textbf{8.4267} & \textbf{6.1997}  & 0.7143         \\
                                                                                                         & \textbf{Z-Intersection} & 0.1771          & 11.7627          & 5.66907           & \textbf{1.0417} & \textbf{0.1700}   & \textbf{11.5800}   & \textbf{4.4850} & 1.0428         \\ \hline \hline
\end{tabular}%
}
\end{table}



\noindent{\textbf{Efficiency towards Various Noise Types.}} Table \ref{tab:noise} illustrates the resilience of the proposed Mahalanobis version of DeepUME~\cite{DeepUME} in the context of point cloud registration tasks subjected to various noise sampling strategies, as outlined in DeepUME~\cite{DeepUME}. Our evaluation encompasses diverse noise types, including Bernoulli, Gaussian, Sampling, and Zero-Intersection, each meticulously detailed within the framework of DeepUME~\cite{DeepUME}. The benchmarking is conducted on the ModelNet40~\cite{modelnet}, FAUST~\cite{FAUST}, and Stanford-3D Repository datasets. While the latter two datasets are exclusively utilized for testing, ModelNet40 is employed for comprehensive evaluation. All training-testing settings\footnote{\href{https://github.com/langnatalie/DeepUME}{https://github.com/langnatalie/DeepUME}} are considered wrt original paper~\cite{DeepUME}.

The outcomes presented in Table \ref{tab:noise} substantiate the superior performance of the proposed MDeepUME over its original counterpart across various datasets and noise types. This observation holds true for a multitude of evaluation metrics, notably including Chamfer distance~\cite{ChamferDistance} (\textbf{CD}), Hausdorff distance~\cite{HausdorffDistance} (\textbf{HD}), and root mean squared error in rotation (\textbf{RMSE(R)}). However, it is worth noting that MDeepUME demonstrates certain limitations, particularly in estimating translation accuracy, as evident from the \textbf{RMSE(t)} values in Table \ref{tab:noise}. This outcome is attributed to DeepUME's intrinsic projection of points into an \(\mathcal{SO}(3)\)-invariant space, wherein the potential of Mahalanobis distance cannot be fully harnessed.

\subsection{Limitations}

While our approach harnesses the prowess of Mahalanobis distance for enhanced feature matching and point cloud registration, it is important to acknowledge its limitations. Mahalanobis distance computation critically depends on the accurate estimation of the covariance matrix. When the covariance matrix is ill-conditioned, the precision may become compromised. Although we mitigate this concern by introducing a small bias term (\(10^{-5}\)) to the diagonal of the covariance matrix. This limitation is evident in noise benchmark evaluation for  DeepUME, where the Mahalanobis version fails in accurately estimating translation parameters. Despite these limitations, our approach significantly contributes to address key challenges in point cloud registration.

\section{Conclusion}

Our paper introduced Mahalanobis k-NN, a statistical framework for improving point cloud registration in learning-based methods. It addresses challenges like feature matching due to variations in point cloud densities. Mahalanobis k-NN leverages local neighborhood distribution for accurate feature extraction, outperforming methods like Flyod-Warshall. It integrates seamlessly into local graph-based point cloud analysis methods like DCP and DeepUME, achieving state-of-the-art performance on benchmark datasets. We also demonstrated that registered point cloud features possess discriminative capabilities, leading to significant improvements in few-shot classification tasks. Mahalanobis k-NN offers surface awareness and outlier robustness. Overall, our proposed method offers a comprehensive and effective solution for point cloud registration, providing superior results, and versatility across various benchmarking test-beds.

{\small
\bibliographystyle{ieee_fullname}
\bibliography{egbib}
}

\end{document}